\documentclass[conference]{IEEEtran}
\usepackage{times}

\usepackage[numbers]{natbib}
\usepackage{multicol}
\usepackage{xcolor}

\usepackage{amsmath,amsthm,amssymb}
\usepackage{graphicx}
\usepackage{caption}
\usepackage{subcaption}
\usepackage[ruled,vlined]{algorithm2e}

\newtheorem{theorem}{Theorem}
\newtheorem{lemma}[theorem]{Lemma}

\makeatletter
\def\blfootnote{\xdef\@thefnmark{}\@footnotetext}
\makeatother

\begin{document}

\title{Human-Robot Team Coordination with Dynamic and Latent Human Task Proficiencies: Scheduling with Learning Curves}


\date{}
\author{\authorblockN{Ruisen Liu$^{\dagger}$}
\authorblockA{Georgia Institute of Technology\\
Atlanta, Georgia 30332--0250\\
Email: ruisenericliu@gatech.edu}
\and
\authorblockN{Manisha Natarajan$^{\dagger}$}
\authorblockA{Georgia Institute of Technology\\
Atlanta, Georgia 30332--0250\\
Email: mnatarajan30@gatech.edu}
\and
\authorblockN{Matthew Gombolay}
\authorblockA{Georgia Institute of Technology\\
Atlanta, Georgia 30332--0250\\
Email: matthew.gombolay@cc.gatech.edu}

}


%

\thanks{Why doesn't this appear?}

\maketitle

\begin{abstract}
As robots become ubiquitous in the workforce, it is essential that human-robot collaboration be both intuitive and adaptive. A robot’s quality improves based on its ability to explicitly reason about the time-varying (i.e. learning curves) and stochastic capabilities of its human counterparts, and adjust the joint workload to improve efficiency while factoring human preferences. We introduce a novel resource coordination algorithm that enables robots to explore the relative strengths and learning abilities of their human teammates, by constructing schedules that are robust to stochastic and time-varying human task performance. We first validate our algorithmic approach using data we collected from a user study (n=20), showing we can quickly generate and evaluate a robust schedule while discovering the latest individual worker proficiency. Second, we conduct a between-subjects experiment (n=90) to validate the efficacy of our coordinating algorithm. Results from the human-subjects experiment indicate that scheduling strategies favoring exploration tend to be beneficial for human-robot collaboration as it improves team fluency (p = 0.0438), while also maximizing team efficiency (p $<$ 0.001).{\let\thefootnote\relax\footnote{{$\dagger$ These authors contributed equally to the work.}}}
  
\end{abstract}

\IEEEpeerreviewmaketitle

\section{Introduction}
\label{Introduction}

Advancements in robotics have opened opportunities for humans and robots to collaborate within joint workspaces, especially in final assembly manufacturing. A human-robot team should be able to leverage its unique strengths to achieve safe, effective, and fluent coordination. However, learning these strengths and ensuring that the team satisfies requisite temporal constraints is challenging. First, humans naturally exhibit static and dynamic forms of variability in task performance. In particular, human task performance (e.g., time to complete a task) represents a static uncertainty in the form of a random variable with a non-negligible variance. Second, human task performance exhibits time variability (i.e., people become more efficient with practice). Third, lack of consideration for human preferences and perceived equality may in the long run, put efficient behavior and fluent coordination at a contradiction \cite{preferences2, preferences1}.

While a human-centered approach may be able to determine ideal task assignments while balancing efficiency and preference, it is also a tedious, manual process. For example, a new assembly line may not reach peak performance until six months, while human workers hone their skills and determine the proper role assignment of each worker. An automated, predictive approach would be especially desirable for seasonal manufacturing demands, where seasonal workers must be quickly yet optimally assigned to fulfill task demands.

Recent advances in scheduling methods for human-robot teams have shown a significant improvement in the ability to dynamically coordinate large-scale teams in final assembly manufacturing \cite{17, 2, 3}. Prior approaches typically rely on assuming deterministic or set-bounded uncertainty throughout each task for coordinating large-scale teams. However, by reasoning myopically, these methods fall short in explicitly reasoning about the uncertainty dynamics in human work that could result in significant productivity gains. These methods are also unable to give guarantees -- probabilistic or otherwise -- of satisfying the upper- and lower-bound temporal constraints in the presence of this uncertainty.
\begin{figure}
\center{\includegraphics[width = .5 \textwidth]{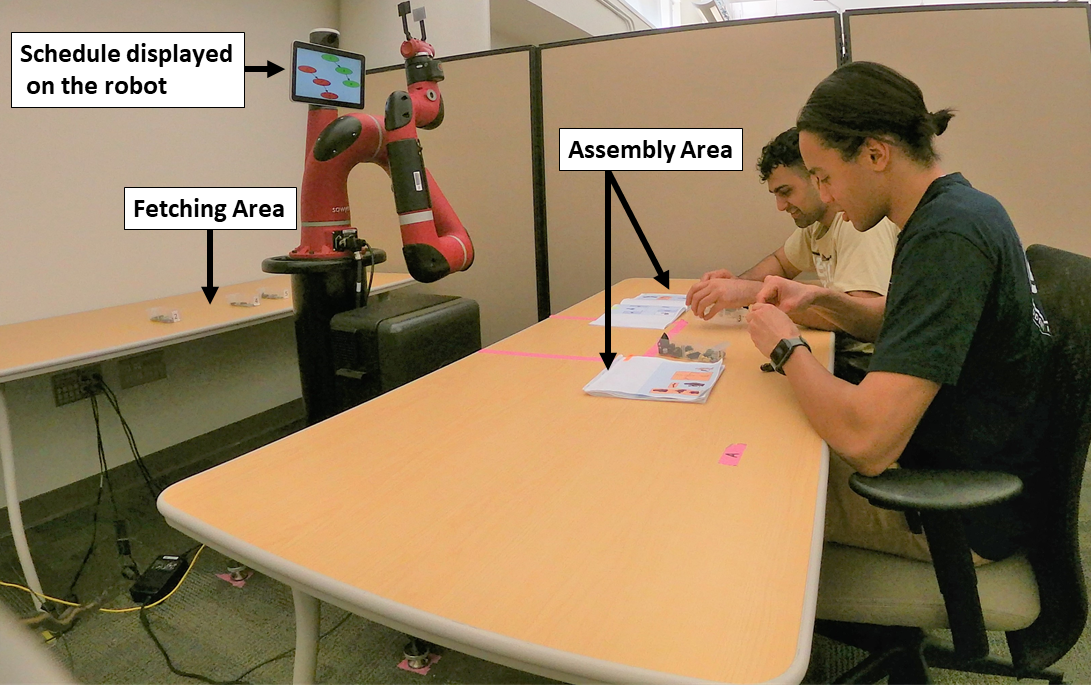}}
  \caption{This figure depicts the setup of our human-subject experiment of a manufacturing analogue environment.}
\label{fig:station} 
\end{figure}
 
In this paper, we formulate a novel, human-aware, computational method for scheduling human-robot teams that synthesizes robust schedules while adapting to the characteristic behavior of participants as they learn skills while following a schedule. In particular, we develop a framework in which a robot reasons about the strengths of its teammates by trying out novel, possibly sub-optimal allocations of work, while still providing a probabilistic guarantee on satisfying the temporal requirements of the manufacturing process. Prior work has shown that switching tasks among team members can lead to greater efficiency and generalizability to new task variants \cite{cross_training, Ramya}. Further, the effectiveness of human-robot collaboration relies heavily on building trust and team fluency among teammates \cite{Trust, 17, CHASKI}. Yet these works have not considered the significantly more challenging problem of robust coordination of human-robot teams under temporal upper- and lower-bound constraints with dynamic, and stochastic task proficiency. Thus, we design a large-scale experiment of 90 human subjects to evaluate trade-offs in exploration-versus-exploitation for maximizing overall team-efficiency, while also trying to infer how trust and team fluency varies across different scheduling approaches.

To accomplish successful human-robot collaboration with improved human-robot trust and team fluency, we develop methods to determine high-quality schedules and adaptions to human performance, with contributions that 1) infer dynamic characteristics about individual human performance and their ability to learn on the task, leading to improved efficiency in task allocation, and 2) enable an efficient computation method for evaluating the robustness of candidate schedules with respect to deadlines, which enables 3) a fast optimization algorithm for iterative improvement of human-robot schedules based on observed progress. We then conduct a full human-subject experiment that yielded statistically significant improvements for team fluency $(p < 0.001)$ and efficiency $(p < 0.05)$ when consistently factoring task diversity over more myopic approaches.

\section{Related Work}
\label{Related Work}

Scheduling for a multi-agent team requires reasoning about temporal, sequential, and spatial constraints \citep{1}. For collaborative human-robot teams, scheduling also requires reasoning about explicit and implicit preferences by human teammates, which in the long-run affects efficiency and willingness to participate \cite{17,  Roy2013, Tsarouchi2017}. In addition, team fluency and trust are important elements to evaluate and maximize during collaboration of human-robot teams \citep{Chao, hoffman, huang_coordination}. An intermediate approach is needed to satisfy these distinct, separate objectives.

Traditional scheduling methods often apply a deterministic approach, using expected or worst-case task duration \citep{2, 3}. In many real-world applications, such an assumption is incompatible with a notion of uncertainty with respect to the duration of each task and human task performance \citep{4}. Proactive approaches seek to reason about uncertainty sets, which define the set of possible values generated by a model \citep{5}. Simple temporal networks, which admit at most one interval constraint on any pair of time intervals, infer about set-bounded uncertainty but do not reason about the uncertainty set itself \citep{6}. 


Optimization-based techniques seek to find a robust, feasible or optimal solution but present a computational challenge, as the computation time of a mixed-integer linear program (MILP) solver grows exponentially with respect to the number of tasks and agents \citep{15}. Robust optimization techniques look to satisfy all constraints based on some measure of robustness \citep{5}. However, linear optimization models for the uncertainty set lead to too conservative bounds, while non-linear models also lead to too expensive computation costs \citep{8,9, 7}. Chance-constrained programming guarantees the probability of meeting a set of constraints to be greater than $1- \epsilon$ \citep{12, 11, 14, 13}. Yet application of existing chance-constrained approaches towards scheduling distributes risk-satisfaction over $O(n)$ individual tasks, resulting in high conservatism. 



Importantly, each optimization technique is missing a built-in process to reason about changing human capabilities, and thus constraints, over time. Some frameworks formulate objectives for multiple additional criteria, such as fatigue, quality, and fairness, but do not explicitly target optimization for the duration of a schedule \cite{mixed_objectives, rl_fairness}.  Kwon and Suh (2014) abstract the challenge of schedule optimization to one of maintaining proactive robot behavior in the face of uncertainty \citep{proactivity}. Gombolay et al. (2015) found that human would prefer to work with robots that factor in their preferences, but not necessarily so if at the expense of team \citep{preferences1}.

Our approach considers the distinct problem of dynamic uncertainty attributed to human ability to improve at tasks through repeated practice. Our approach accomplishes the following novel points:

\begin{itemize} 
    \item We apply risk allocation to deadline rather than task constraints as an extension of chance-constrained programming, reducing the distribution of risk and minimizing the conservatism of bounds.
    \item We provide parametric models and updates to explicitly reason about individual human task completion ability.
    \item We conduct a large-scale study (n=90) to investigate human preferences within the situational context of human-robot teams under temporal constraints and heterogeneous task proficiency.
    \item We explore how our algorithm can maximize trust and team fluency without sacrificing efficiency. 
\end{itemize}

To the best of our knowledge, our approach is the first to reason explicitly about risk allocation with respect to deadlines while also tackling dynamic uncertainty and projecting human learning capability.

\section{Motivation}
\label{Motivation}

\subsection{Problem Formulation}
\label{formulation}

We formulate our objective function to minimize the maximum amount of work any agent performs (makespan) while penalizing for a measure with respect to the average number of untried agent-to-task assignment combinations (entropy). Along the way, we explicitly reason about each human and robot's individual strengths and weaknesses in task completion and ensure that natural variability in human performance is accounted for while aiming to meet deadlines. Our objective function is subject to time-based constraints related to probabilistic task completion times, waits, upper-bounds (deadlines) and lower-bounds (temporal constraints).

Let $n_t$ denote the number of tasks and $n_a$ denote the number of agents. Let $\tau_i $ be a task completed $N_{\tau_i}$ times, and $\tau_{i_n}$ denotes the $n^{th}$ iteration of said task. Let $ A_{\tau_{i_n}}^a \in \{ 0, 1 \} $ be a binary variable indicating the assignment of agent $a \in A$ to the $n^{th}$ iteration of task $ \tau_i$. Let $r_{\tau_i}^a$ be the number of repetitions that agent $a$ has completed for task $\tau_i $. Let $\hat{c}_{i_n}^a (x) $ be a function estimating the expected duration of agent $a$ completing iteration $x$ of task $ \tau_i$.

\begin{gather}
\min z =  z_1 + \lambda z_2\label{eq1} \\
z_1 = \max_{\tau_{i_n} \in \tau}  f_{\tau_{i_n}} \label{z1} \\ 
z_2 = \frac{1}{n_t n_a} \sum_{\tau_i} \left[  \sum_{a} | \frac{1}{n_{a}} \sum_{a} r_{\tau_i}^a - r_{\tau_i}^a | \right] \label{z2} \\
r_{\tau_i}^a = \sum_n A_{\tau_{i_n}}^a , \forall \ \tau_{i_n}  \in \tau, \forall a \in A \label{eq_mcg_6} \\
\sum_{a \in A} A_{\tau_{i_n}}^a   = 1 , \forall \ \tau_{i_n}  \in \tau \label{eq2}\\ 
\left(s_{\tau_{j_m}} - f_{\tau_{i_n}}\right)x_{\tau_{i_n}, \tau_{j_m} }A_{\tau_{i_n}}^a A_{\tau_{j_m}}^a \geq 0, \nonumber \\\forall {\tau_{i_n}, \tau_{j_m}} \in \tau, \forall a \in A \label{eq4} \\ 
x_{\tau_{i_n}, \tau_{j_m}} + x_{\tau_{j_m}, \tau_{i_n}} = 1, \forall  \tau_{i_n}, \tau_{j_m} \in \tau  \label{eq_mcg_5} \\
s_{\tau_{j_m}} - f_{\tau_{i_n}} \geq W_{\tau_{i_n}, \tau_{j_m}} \forall {\tau_{i_n}, \tau_{j_m}} \in \tau |  W_{\tau_{i_n}, \tau_{j_m}} \in TC
\label{eq6} \\
ub_{\tau_{i_n}} \geq f_{\tau_{i_n}} - s_{\tau_{i_n}} \geq lb_{\tau_{i_n}} \label{eq_mcg_1} , \forall  \tau_{i_n}  \in \tau \\
D_{\tau_{i_n}, \tau_{j_m}}^{rel} - ( (f_{\tau_{j_m}} - s_{\tau_{i_n}}) + \left[ \Phi^{-1} ( 1 - \epsilon_D)   \sigma_{D_{\tau_{i_n}}} \right]) \geq 0 , \nonumber\\ \forall {\tau_{i_n}, \tau_{j_m}} \in \tau | \ \exists D_{\tau_{i_n}, \tau_{j_m}}^{rel} \in TC, \sum_{D \in TC } \epsilon_D = \epsilon
\label{eq7}\\
D_{\tau_{i_n}}^{abs} -  (f_{\tau_{i_n}} + \left[ \Phi^{-1} ( 1 - \epsilon_D)   \sigma_{D_{\tau_{i_n}}} \right]) \geq 0,\nonumber \\ \forall \tau_{i_n},  \in \tau | \ \exists D_{\tau_{i_n}}^{abs} \in TC \land \sum_{D \in TC } \epsilon_D = \epsilon \label{eq8}
\end{gather}



We define our objective function in Eq. \ref{eq1} as a linear combination of minimizing the makespan in Eq. \ref{z1} and entropy in Eq. \ref{z2}. Eq. \ref{eq2} requires all tasks $\tau_{i_n}$ to be assigned to an agent. Eqs. \ref{eq4} and \ref{eq_mcg_5} require each agent to complete one task at a time. Eq. \ref{eq6} requires a minimum of $W_{\tau_{i_n}, \tau_{j_m}}$ between the end of task $\tau_{i_n}$ and the start of task $\tau_{j_m}$. Eq. \ref{eq_mcg_1} sets a potential hard upper- and lower-bound for the duration of task $\tau_{i_n}$.


To optimize subject to chance constraints for our uncertainty in task completion time, we enforce a deadline-based risk allocation rather than a task-based risk allocation. This reduces the risk allocation spread of $\epsilon$ from $O(n)$ tasks to $O(D)$ deadlines. We express the resulting risk allocation with Eqs. \ref{eq7} and \ref{eq8}. Eq. \ref{eq7} requires the likelihood of completing relative deadline $D_{\tau_{i_n}, \tau_{j_m}}^{rel}$ between the start of task $\tau_{i_n}$ and the end of task $\tau_{j_m}$ be greater than $1 - \epsilon_D$, where $\Phi $ denotes the cumulative normal distribution function. Eq. \ref{eq8} requires that the likelihood of completing each absolute deadline $D_{\tau_{i_n}}^{abs}$ for task $\tau_{i_n}$ be greater than $1 - \epsilon_D$. 


\subsection{Technical Challenges in Direct Optimization}
\label{optimization_challenges}

There are two primary challenges that prohibit a MILP solver from directly addressing the overall problem using this formulation.

\begin{enumerate}

\item \textbf{First, this formulation alone does not provide any adaptive reasoning about individual agent capabilities to complete tasks.} Consequently, maintaining the same schedule misses opportunities to optimize on inter-agent skill capabilities.

\item \textbf{Second, the variable} $\mathbf{\sigma_{D_{\tau_{i_n}}}}$\textbf{in Eqs. \ref{eq7} and \ref{eq8} is under-defined, and dependent on schedule assignments.} The variable $\sigma_{D_{\tau_{i_n}}}$ relates to the probabilistic distribution of times for completing task $\tau_{i_n}$. As such, you cannot compute $\sigma_{D_{\tau_{i_n}}}$until a full assignment has been made for all tasks and task orders, \{$ A_{\tau_{i_n}}^a , x_{\tau_{i_n}, \tau_{j_m} }$\}, associated with requisite tasks leading up to the task deadline. This is also the case for computing the probabilistic distribution of the makespan. The evaluative process becomes computationally expensive due to operations that render each deadline-specific probability distribution non-parametric.  We fully elaborate on this cost and our alternative solution in Subsection \ref{fastapproxalg}.

\end{enumerate}

\noindent Below, we detail our methods to tackle these challenges for our dynamic scheduling environment.

\section{Coordination Algorithm}
\label{algorithm_overview}

To generate an adaptive method that improves upon optimization of static task completion estimates, we seek an approach that evaluates fixed robot task completion times with dynamic, improving human task completion times. Without being able to directly use a MILP solver, we require alternative search processes that will evaluate a large number of candidate schedules. Given the computational expense to evaluate a single schedule, we must derive and leverage a sufficiently fast upper-bound evaluation method. In addition, we need a reliable parametric update to the expected agent-specific task duration duration as we observe agents over time.

\begin{figure}[!htb]
\center{\includegraphics[width = 0.4\textwidth]{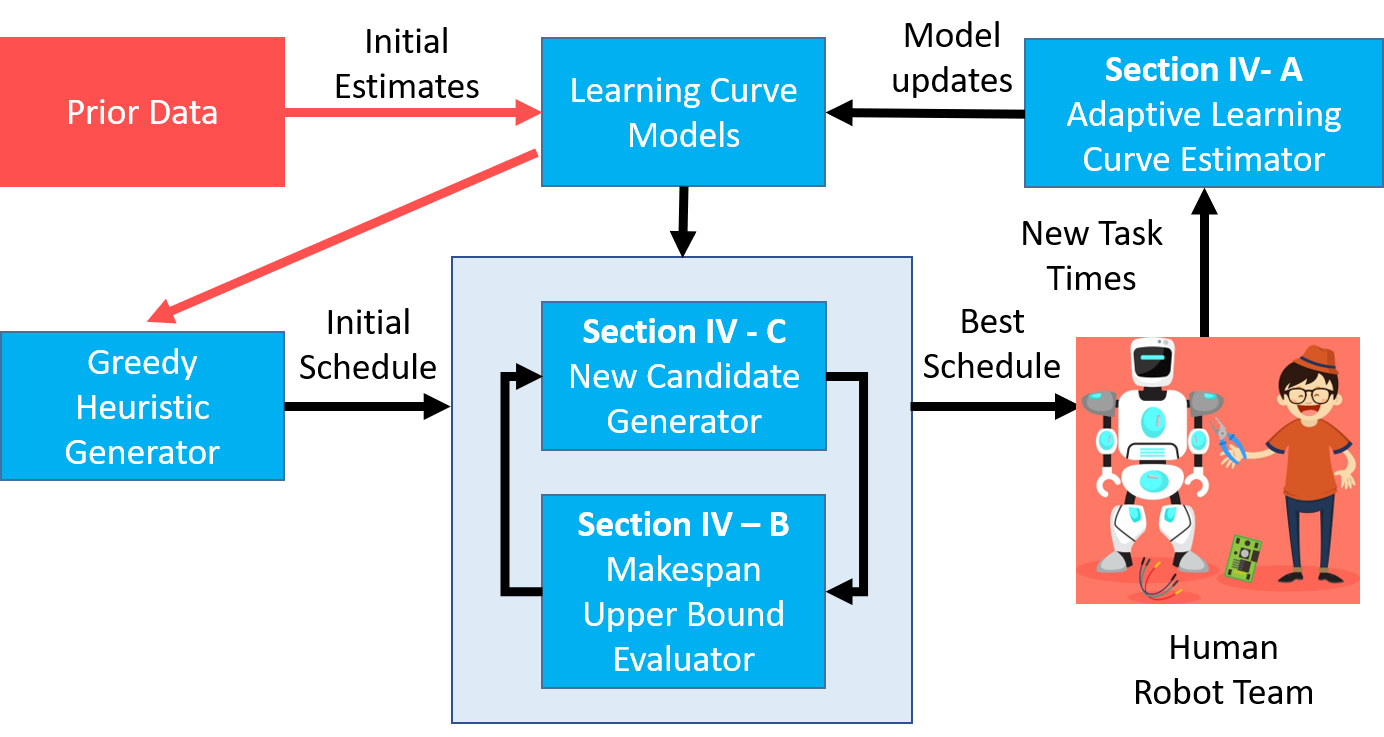}}
\caption{This figure depicts our algorithm's architecture.}
\label{fig:flowchart}
\end{figure}

In the following subsections, we first validate two parametric assumptions in Subsection \ref{assumptions} that are used to parameterize human task completion times. Then, in Subsection \ref{fastapproxalg} we offer a fast upper-bound calculator as a substitute for the objective value, to achieve fast estimation of the robustness and makespan of any candidate schedule. Subsequently, we reason about agent capabilities by computing an update to the estimate for a person's likely future task completion time in Subsection \ref{lcupdate}. We utilize these two methods to facilitate an iterative search process for finding a schedule that balances minimizing the time needed for completion and exploring new task-agent assignment pairs, robust to deadlines. Our approach is depicted in Fig.~\ref{fig:flowchart}. 

We generate an initial schedule using a simple heuristic. Then, we used population-based optimization methods (Subsection \ref{search_methods}) for improving the schedule. This evaluation process is made timely by the fast upper-bound evaluation detailed in Subsection \ref{fastapproxalg}.  Once a new optimal schedule is selected, we pass the schedule to the human-robot team for completion. The duration of each executed task is recorded and used to update the parameters of each corresponding assigned agent's learning curve, as detailed in Subsection \ref{lcupdate}. The learning curve is then used to re-project estimations for the expected duration of future tasks. We repeat as long as scheduling is needed. Our approach is also depicted in Algorithm 1 in our supplementary document. 

\subsection{Algorithm Assumptions and Validation}
\label{assumptions}

We make two assumptions about human task completion capabilities which we validate through a user study. First we assume humans learn over time and that their learning curve follows an exponential function over the course of multiple iterations, as shown historically \citep{learning_curve}. Second, we assume that for a population of people completing an assembly task, the population distribution of task times is approximately normal. 

    To confirm the first and validate the second assumption, we obtained real-world data for learning curves by recording the time taken by different subjects to assemble LEGO kits of varying difficulty. Twenty participants were assigned to complete six different tasks for five iterations each. We test the normality assumption for task completion times on all the six tasks for each iteration using Shapiro-Wilk's test with Bonferroni correction (as we are testing for 30 combinations). Results from Shapiro-Wilk's tests find that we do not reject the null hypothesis that the distribution is normally distributed at the $\alpha = 0.05$ confidence level for 29/30 task-iteration combinations, indicating that 97\% of the data comply with our assumption. Furthermore, the average time taken per task decreased following an exponential function as a function of the number of iterations, as suggested by literature \citep{learning_curve}. As such, we are confident in our two model assumptions in generating simulated schedules and tasks.

\subsection{Fast Upper-Bound for Schedule Robustness}
\label{fastapproxalg}

We develop an upper-bound to address the challenge of under-defined chance constraints in Eqs. \ref{eq7} and \ref{eq8} and for evaluating the makespan in Eq \ref{z1}. Consider the following computation requirements for evaluating a candidate schedule's makespan and deadline robustness. We define the schedule as robust if it is possible to complete all tasks by the desired deadlines with probability $ (1- \epsilon)$. An approach to distributing risk across deadline success ($S_i$) can be obtained via Boole's inequality, as shown in Eq. \ref{boole}, which enables expression of the joint chance constraints of $n$ deadlines as shown in Eq. \ref{jointcc}.
\begin{gather}
Pr[\underset{i=1}{\overset{n}{\cup}} S_i] \leq \sum_{i=1}^n Pr[S_i],\label{boole} \\
Pr(S_i) \geq (1 - \epsilon_i), \sum_{i=1}^n \epsilon_i \leq \epsilon \label{jointcc}
\end{gather}

Boole's inequality allows us to select the appropriate $\epsilon_i$ for each deadline in Eqs. \ref{eq7} and \ref{eq8} given an overall schedule $\epsilon $. To compute the robustness given that $\epsilon_i$, we need to evaluate the probabilistic distribution for the completion of the last task associated with that deadline. Thus, it is necessary to traverse the directed acyclic graph (DAG) of the schedule and its associated task sequence to ascertain the expected completion time of each task. As it is unwieldy to repeatedly reference $A_{\tau_{i_n}}^a $ for task assignment, let the $i^{th}$ task assigned to agent $a$ in a single schedule be denoted as $\tau_{i,a}$. Let $X_{C_{\tau_{i,a}}}$ be a continuous, independent random variable denoting duration of task $\tau_{i,a}$, and let $X_{\tau_{i,a}}$ be the finish time of task $\tau_{i,a}$ during the schedule. We can express $X_{\tau_{i,a}}$ as a function of the time elapsed before the start of the task due to preconditions and the duration of the task as shown in Eq. \ref{deadlineEQ}.

\begin{equation}
    X_{\tau_{i,a}} = max ( X_{\tau_{i,a}}^{precon} + w_i^{precon}  , X_{\tau_{(i-1, a}}) + X_{C_{\tau_{i,a}}}
\label{deadlineEQ}
\end{equation}

In Eq. \ref{deadlineEQ}, $ X_{\tau_{i-1,a}}$ is the end time of the previous task completed by agent $a$, and $X_{\tau_{i,a}}^{precon} + w_i^{precon}$ denotes the satisfaction of a task precondition and wait constraint. This equation requires two computations: a maximization (prior precondition tasks in parallel) and an addition (new task in series). Generally, the maximization of a set of random variables $Y = max \{ X_i \}_{i = 1, 2 .. n}$ yields a resulting probability distribution $ Y \sim f_y(y)$ which can be expressed by Eq. \ref{maxofprobs}.
\begin{equation}
    f_y(y) =  \sum_{i=1}^n f_{x_i} (y) \overset{n}{\underset{j=1, j \neq i}{ \Pi}} F_{x_j} (y)
\label{maxofprobs}
\end{equation}
The subsequent probability distribution resulting from the sum of two random variables $Z = Y + X$, $Z \sim f_z(z)$ can be expressed by Eq. \ref{sumofprobs}.
\begin{equation}
    f_z(z) = \int_{-\infty}^{\infty} f_x (x) f_y (z - x ) dx 
\label{sumofprobs} 
\end{equation}

In summary, by computing a maximization and then a summation operation for each task, we can obtain the probabilistic distribution for each task's finish time via Eq. \ref{deadlineEQ}. We can check if the distribution of the final task before each deadline satisfies the $\epsilon$ allocated via Boole's inequality via Eq. \ref{jointcc}. Finally, we can compute the makespan's distribution by evaluating the probability distributions' maximum associated with each agent's final task. However, we note that consecutive multiplication and integration computations quickly become expensive, as there exists no parametric form to represent the output probability distribution. 

However, since we've shown that a task duration is normally distributed, we can achieve a quick approximation for Eq. \ref{deadlineEQ}, such that the iterative computation mentioned in Subsection \ref{optimization_challenges} is scalable. We validate this assumption of normality in Subsection \ref{simu_fidelity}. The sum of normally distributed random variables computed via Eq. \ref{sumofprobs} has a quick closed-form solution. If $X \sim N(\mu_x,\sigma_x^2)$ and $ Y \sim  N(\mu_y,\sigma_y^2)$, then $Z$ can be parametrized by $ Z \sim N ( \mu_x + \mu_y, \sigma_x^2 + \sigma_y^2)$. However, the max of normally distributed random variables given by Eq. \ref{maxofprobs} does not have a closed-form solution, unless $\exists y \in \mathbb{R} \  s.t. \ F_{x_i} (y) \approx 0,   F_{x_j} (y) \approx 1 \ for \ i \in N, j \neq i$. That is, that the rightmost normal distribution does not ``overlap'' with the distributions. Otherwise, the expression can only be evaluated via integration for every task in the schedule. 

As the computation cost scales exponentially in time with the number of tasks and schedules to evaluate, this computation still becomes prohibitively expensive for problems with a large number of tasks. Instead, we search for a fast approximating Gaussian distributed upper-bound for Eq. \ref{maxofprobs} and introduce a small amount of conservatism in return for a scalable evaluation. Formally, for $Y = max \{ X_i \}_{i = 1, 2 .. n}$, we search for a distribution $f_g(y)$, such that Eq. \ref{lsearch} holds.
\begin{equation}
F_g(y) \leq F_y(y) \ \forall y \in [0, \infty]
\label{lsearch}
\end{equation}


\begin{lemma}
There exists a Gaussian cumulative probability distribution $g \sim N(\mu, \sigma)$ s.t. Eq. \ref{lsearch} is true with probability $1-\epsilon$, for arbitrarily small $\epsilon$.
\end{lemma}

Proof: For $F_y(y), \exists \ b \ s.t. \ (1-\epsilon) \leq F_y(b)$. For an arbitrary $\sigma, \exists \ N(\mu, \sigma), \ s.t. \ \mu - k\sigma \geq b, F_N(\mu - k\sigma) \leq \epsilon$,  for $\mu, k \in [0, \infty]$. Since $\epsilon$ is arbitrarily small, then $F_n (\mu - k\sigma) \leq \epsilon << 1 - \epsilon \leq F_y(b) \leq F_y (\mu - k\sigma) $.

Through experimentation, we found a good initial guess for upper-bound distribution is $f(g)(y) \sim \mathbb{N}(\mu = max(\mu_{x_i}), \sigma = mean(\sigma_{x_i})$, which is equivalent to selecting the mean of rightmost distribution. With the distribution average, we can then perform line search for incremental shift of $\mu$ and $\sigma$ by step sizes $\alpha$ and $\beta$, respectively, until Eq. \ref{lsearch} is satisfied. 
If $\alpha$ is scaled with respect to $\sigma/c$, the maximum steps needed is a single-digit multiplier of $c$. Checking the exit condition at each step can be inexpensive, since although $f_y (y)$ has no closed-form solutions, $F_{x_i}(y)$ in $F_y (y) = \overset{n}{\underset{i}{ \Pi}} \ F_{x_i} (y)$ can be evaluated with a single look-up. Due to Gaussian distribution’s symmetric properties, the number of evaluation points to produce a tight bound can be empirically as few as 12 points spaced uniformly within $[\mu-3\sigma, \mu + 3\sigma]$.

\subsection{Learning Curve Update Algorithm}
\label{lcupdate}

We address the other challenge of modeling humans’ learning capability on a particular task by parametrically reasoning about a human's hidden dynamic model. We can expect to model the learning curve as an exponential function of generic form $y = c + ke^{-\beta i}$, where $c, k, b $ are parameters and $i$ is the repetition number \citep{learning_curve}. However, humans exhibit variance in task completion time. Since a person learns at each iteration, it is hard to attribute a change in task completion time to learning or to noise. While we can compute a model prior by looking at population averages, we need a way to tailor a model update for each individual. We design our learning curve model update as an adaptive Kalman filter detailed in Fig. 1 from \citep{adaptivekalman}, where our state vector is composed of learning curve parameters and our observation consists of observed task duration. Extending the Kalman analogy, we model each individual's learning curve parameters as a hidden but static state, with the best initial guess as the population average. After each task is completed, we obtain a new observation and update our parameters. 
Our primary challenge comes from determining analogous initial estimates for the state, prior covariance and covariance noise matrices, $P, Q, R$, and the corresponding update rate $\alpha$. To find a suitable estimate, we bootstrap by repeatedly sub-sampling the population prior to computing the initial covariance matrix $P$ and scale $Q$ and $R$ in proportion to $P$. To determine the ideal update rate for $Q$ and $R$, we run repeated simulations to sweep through range $[0, 1]$ in intervals of $0.25$ and find $\alpha = 0.9$ to be on average the best update rate.

    

\subsection{Population Based Optimization} 
\label{search_methods}

MILPs are often solved via a branch-and-bound solver, which require non-trivial upper- and lower-bounds for pruning sub-trees, and a branching order for binary variables. While we develop an upper-bound formulation in Subsection \ref{fastapproxalg}, we note that finding a non-trivial lower bound that will accelerate the efficiency of a branch-and-bound approach by frequently pruning sub-trees by exceeding this upper-bound is not likely given a wide domain on uncertainty sets. Alternatively, we can search via population-based optimization. We propose to generate an initial best schedule using some known heuristic, such as EDF, and search via iterative mutation by swapping task allocations and order starting from this schedule candidate. We employ an evolutionary-based approach to sequentially generate new candidates, keeping a percentile of best candidate schedules as defined by our objective function. Candidate schedules are also evaluated for robustness to deadlines; those that are not robust are eliminated. After successive rounds of evolution, the best schedule is selected for implementation.

\section{Empirical Algorithmic Evaluation}

In this section, we simulate complex scheduling problems to confirm the general ability of our individual algorithm components to provide a fast, tight upper-bound and reason about individual capabilities. 

\subsection{Simulation Generation}
\label{simu_fidelity}

To generate random synthetic stochastic scheduling problems, for each agent and task, we generate each $c, k, b$ parameters of that agent's learning curve by aggregating three components, which respectively represent the task, agent, and joint task-agent contributions to that parameter. Each component is sampled from a reasonably proportioned normal or uniform distribution. We repeat to generate a proportionally scaled hidden variance at every iteration, representing the agent's potential to deviate from their learning curve.

We then tune some parameters to make the scheduling problem challenging with respect to time constraints. The total time allocated  is set to $\frac{\mu + 3 \sigma}{n_a}$, where $\mu, \sigma$ are the total expected time of all tasks and its corresponding variance and $n_a$ is the number of agents. On average, 1/5th of all tasks have a corresponding deadline and each task a weighted probability of having zero to three preconditions. If the task has a precondition, there is a 50\% chance of  an associated wait period, sampled uniformly with a finite range $[a, b]$. We choose to generate scheduling problems with less than 100 tasks and up to three agents, as literature and online MILP solvers indicate that solutions are not tractable past the trade-off curve of those milestones \citep{15}.
\begin{figure}[t]
 \center{\includegraphics[width = .475 \textwidth]{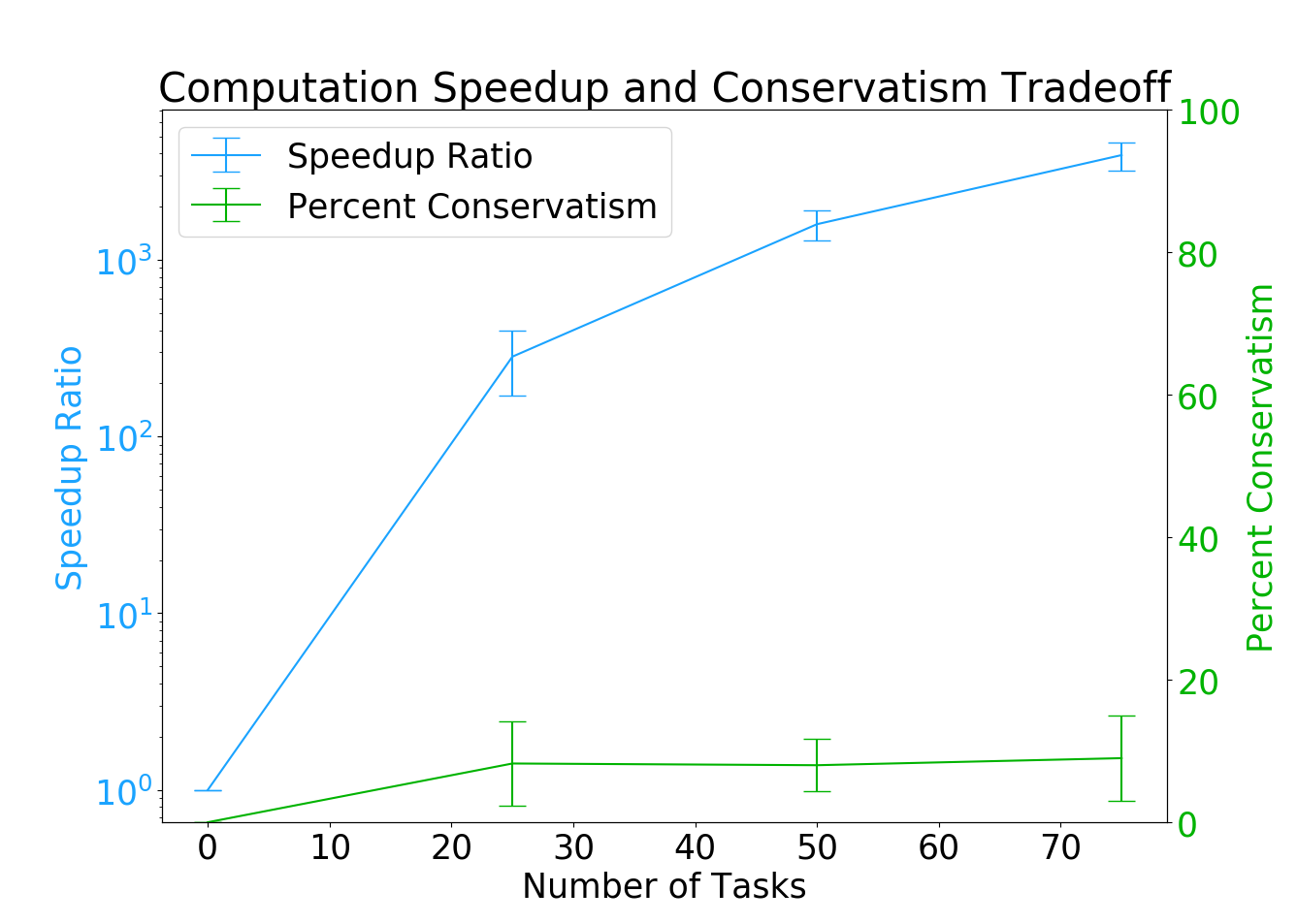}}
  \caption{This figure depicts the trade-offs in employing numerical quadrature for a probability distribution over satisfying temporal constraints vs.~our upperbound.}
\label{fig:Gauss_approx} 
\end{figure}

\subsection{Computation Time Advantage for Robustness Evaluation}
\label{computation_costs_results} 

We constructed feasible schedules using an earliest deadline first heuristic for three representative agents. We then evaluated the average computation time costs for deriving an accurate probabilistic distribution for the makespan of the schedule using both numerical estimation and Gaussian upper-bound approximation given $n$ tasks. While computing a Gaussian approximation always took less than 0.1s for 25, 50, and 75 tasks, the corresponding numerical approximation took 5.56 $\pm$ 1.53s , 51.69 $\pm$ 2.53s, 201.14 $\pm$ 57.75, respectively at a 95\% confidence level. We show the speedup ratio in Fig. \ref{fig:Gauss_approx}. As such, the computation requirements make it impractical to evaluate a large number of schedule candidates via numerical approximation. For a schedule robustness requirement of 95\% probability of success, the average amount of conservatism (percent added time) introduced by using our approximation method was 8.44 $\pm$ 10.1\%  at a 95\% confidence level.

\subsection{Efficacy of Learning Curve Update}
\label{kalman_results}

\begin{figure}[t]
\center{\includegraphics[width = .29 \textwidth]{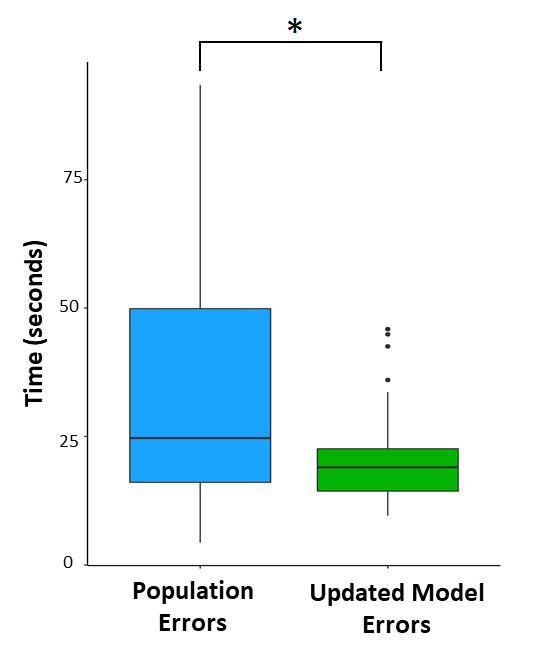}}
  \caption{This figure depicts how our adaptive model achieves a statistically significant improvement in inferring the capabilities of the robot's human teammates ($p < 0.01$).}
\label{fig:Kalman_update_new} 
\end{figure}

To demonstrate our algorithm’s effectiveness in discovering individual agent’s learning curve, we generated 50 instances where an initial learning curve was estimated from a prior of 50 agents sampled for 20 iterations. A new agent was then introduced with unknown parameter, and the total error of each model with respect to the true parameters was aggregated across iterations. The population model showed a median total error of 24.7s (10.1\%). The updated model showed a median total error of 19.0s (7.8\%) . Crucially, update model removed many outliers found in the population data. The average improvement across trials was measured at 13.7s, representing a 60\% increase in prediction accuracy by changing from the population model to the learned individual model as show in Figure \ref{fig:Kalman_update_new}. While these tasks are relatively short, when scaled up to factory-level, this improvement would have a strong impact.

In the following section we apply the entirety of our algorithm in a human subjects study, and validate the efficacy of our approach in factoring in trust and team fluency, while also maximizing team efficiency.

\section{Human Subjects Study}
\label{sec:human_subject_study}

 We conducted a between-subject study (n=90) to gain insight into how a robotic scheduler should balance between exploration and exploitation strategies for a human-robot team in an analogous manufacturing setting. This study examined trends in schedule makespan, subject trust, task fluency, and perceived task difficulty for different scheduling approaches in a LEGO assembly task. To facilitate the schedule updates, we obtained an initial parametric estimate for the learning curve of each task in a pilot study. We collected 20 trials of five completed repetitions for each of the six LEGO tasks.

We investigated three different scheduling approaches in this study (exploitation, explore-exploit, and annealed explore-exploit), balancing participants across conditions.

\begin{itemize}
    \item \textbf{Exploitation:} The algorithm seeks to minimize the schedule makespan based on prior task completion times from the pilot study. We set $\lambda = 0$ in Eq. \ref{eq1}.

    \item \textbf{Explore-Exploit:} In addition to minimizing the makespan, the algorithm is weighted towards considering novel assignments during the entire session by setting $\lambda = 50$, which weights diversity of task assignment in proportion to a schedule that has an average makespan of 360 seconds.
   
    \item \textbf{Annealed Explore-Exploit:} The algorithm is weighted towards considering novel assignments in the first half of the session but the only seeks to minimize the makespan during the second half of the session.
    
\end{itemize}

We received approval from our Institute's ethics committee to conduct human-subject experimentation. The study consists of two participants collaborating with a robot scheduler in one session. The subjects participate in five rounds of LEGO assembling tasks, each with three tasks per participant in each round. The tasks are designed to be about two minutes each. The robot is controlled by the investigator and is assigned to fetch bins before a human teammate completed their assigned task. Both participants are asked to record the task completion times for each task.


\subsection{Hypotheses}

\textbf{H1} \textit{Schedule Makespan Improvement: Explore-Exploit will have greater improvement in schedule makespan, than pure exploitation or annealed explore-exploit.} Past studies have shown that exposing members to different tasks can lead to increased overall efficiency and generalization to new task variations in small-scale teams \cite{cross_training}.

As explore-exploit accounts for greater entropy, we hypothesize that it can lead to greater improvement in task completion times. Schedule makespan improvement is calculated as the percentage improvement in completion times of the last round with respect to the first round.

\textbf{H2} \textit{Robot Trust: Inclusion of some exploration in scheduling will engender greater trust.} Exploration favors novel assignments to the participants. We hypothesize that explore-exploit and annealed explore-exploit can engender greater trust as the participants can comprehend how the robot is trying to optimize their schedule. Robot trust is assessed using a 7-point Likert scale (1 indicates ``Strongly Disagree'' and 7 indicates ``Strongly Agree'') survey taken from \cite{hoffman}.

\textbf{H3} \textit{Team Fluency: Perception of team fluency will increase in conditions with exploration in schedule assignments.}  Studies in the past have shown that switching tasks between teammates can lead to developing a shared mental model of the task, and thereby the teammates can better anticipate each other's actions \cite{cross_training}. Hence, we hypothesize that exploration-based strategies can lead to greater team fluency in human-robot teams. Team fluency is calculated using a 7-point Likert scale of the working alliance questionnaire for human-robot teams from \cite{hoffman}.

\textbf{H4} \textit{Perceived Difficulty: Participants' perception of task difficulty will not depend on the scheduling strategy.} As all the six tasks are designed to be approximately of the same duration, we hypothesize that total task difficulty will not depend on the scheduling approach. Task difficulty is assessed using a 7-point Likert scale of the task difficulty questionnaire from \cite{Chao}.

\subsection{Procedure}

At the start of the experiment, both human teammates read and sign a consent form explaining the study duration, and their compensation for participation. The setup consists of  the robot being placed between two tables, with LEGO kits on the first table in bins and assembly stations marked on the second table. Both human participants are asked to sit at the second table opposite the robot as shown in Fig. \ref{fig:station}. 

Our experiment consists of two types of tasks - fetching part kits and assembling part kits. The robot is capable of only fetching part bins, while the human workers assemble the tasks. The fetching task involves inspecting the bins and transporting them to the assembly area. The setup is designed to mimic a manufacturing final assembly, which involves regular inspection of parts when transported from one zone to another.

A scheduling approach is chosen at random for every session. The session begins with the robot announcing the initial schedule as ``Hi. We will be investigating a different schedule approach today in this session.'' The first round starts with an initial schedule where the tasks are distributed almost equally (based on prior times collected) between the two subjects to minimize the makespan. At the start of each round, the robot displays the schedule in a graph that shows the assigned tasks to each subject. The robot then started fetching bins per the schedule and delivered them to the assembly area. The two subjects started each round of tasks together and timed themselves for the number of seconds they took to complete each task. The investigator checked to make sure that both the subjects were timing themselves appropriately. 

At the end of each round, the subjects hand over the times to the operator, which are then fed into the algorithm to come up with a different schedule. The robot then comments again on whether the tasks were going to be swapped (explore condition) or reassigned based on time taken on the previous round (exploit condition). If the previous round was using an exploitation-based update, the robot would say ``Based on the time you took in the previous round, I think this should be the new schedule.'' If the previous round was using an exploration-based update, the robot would say, ``We're going to try to mix it up and see how people do. Here's the new schedule.''

At the end of five rounds, the operator concludes the session and the participants are then asked to fill a 7-point Likert scale questionnaire to measure team fluency, task difficulty and trust in the robot.  We provide our questionnaire in the supplementary documents.

\subsection{Study Results}
\label{sec:exp_result}
Statistical analysis of all hypotheses are reported here. The significance level $\alpha$ = 0.05. For analyzing H2-H4, we consider a linear mixed effects model with random effects. The session number (indicates which session the subject participated in, with each session having two subjects) is taken as a random effect in this model. The independent variables considered include - scheduling approach (categorical variable with three levels), total time taken by the subject to complete all five rounds of tasks, and difference in completion times between participants. We consider a simple model for the analysis of H1 as makespan improvement is dependent on the completion times.

\subsubsection{Analysis of H1 - Schedule Makespan Improvement}
\begin{figure}

\center{\includegraphics[width = 0.50\textwidth]{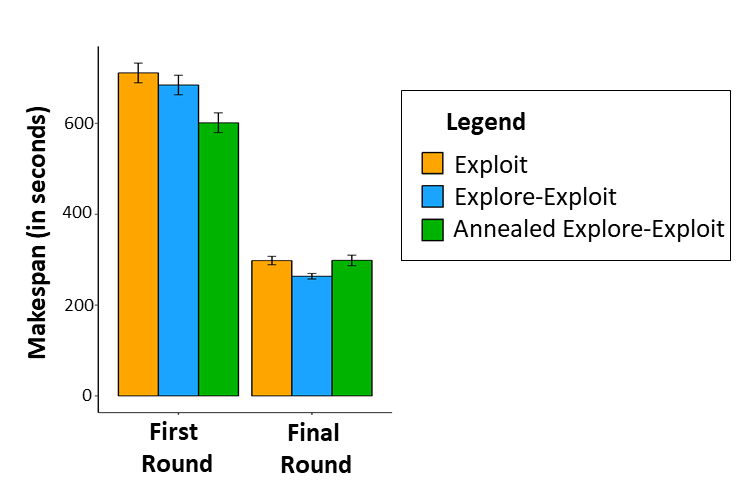}}
  \caption{Average makespan by round. Explore-Exploit results in the shortest net makespan. }
\label{fig:makespan} 
\end{figure}

In the analysis of H1, we consider a simple linear model with makespan improvement as the dependent variable and scheduling strategy as the independent variable. As this model fails the normality of residuals assumption of ANOVA (Shapiro-Wilk's test shows $p = 0.002$), we resorted to use a non-parametric test, namely Kruskal-Wallis for our analysis. Results from Kruskal-Wallis shows that the schedule makespan improvement is highly significant on the type of scheduling strategy ($p < 0.001$). Post-hoc analysis with Dunn's test shows both exploit and explore-exploit strategies to be significantly better than annealed explore-exploit with p-values as 0.024 and 0.00037 respectively. Trends indicate that makespan improvement is the highest for explore-exploit, then pure exploitation and annealed explore-exploit as shown in Fig. \ref{fig:makespan}. We also observe that the largest net improvement in explore-exploit.

\subsubsection{Analysis of H2 - Robot Trust}
Robot trust, assessed using a 7-point Likert scale survey taken from \cite{hoffman}, is considered the dependent variable in the linear mixed effects model with task completion times and strategy as discussed above. After verifying the model follows the assumptions of ANOVA using Shapiro-Wilk's ($p = 0.4783$) for normality of residuals and Levene's Test ($p = 0.052$) for homoscedasticity, we perform an ANOVA. We obtain marginal significance for the scheduling strategy ($p = 0.08603$). Although we fail to reject the null hypothesis, the trends in robot trust seem to be promising with annealed explore-exploit garnering the highest trust, closely followed by explore-exploit. Trust in exploitation is much lower than the former two approaches.

\subsubsection{Analysis of H3 - Team Fluency}
Team fluency, calculated using a 7-point Likert scale of working alliance questionnaire  from \cite{hoffman}, We consider team fluency as the dependent variable in the above mentioned linear mixed effects model. We confirm that this model adheres to the assumptions of ANOVA by Shapiro-Wilk's Normality test ($p = 0.7418$) and Levene's test for homoscedasticity ($p = 0.20963$). We obtain statistical significance for scheduling strategy ($p = 0.0438$) after ANOVA. Post-Hoc analysis with Tukey-HSD shows significance for explore and exploit strategies ($p = 0.0227$).

\begin{figure}

\center{\includegraphics[width = 4.2cm, height = 6.25cm]{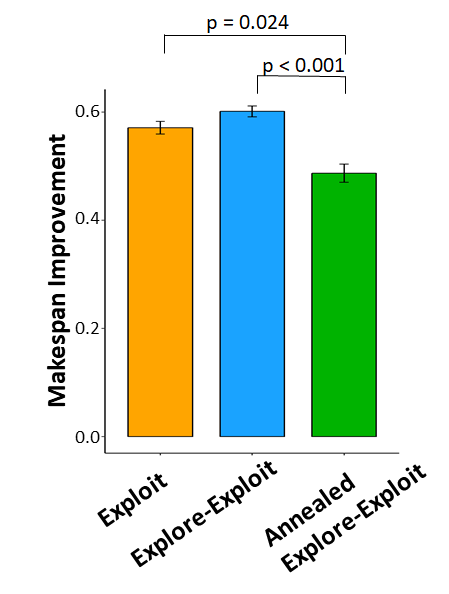}}
\includegraphics[width = 4.2cm, height = 6.25cm]{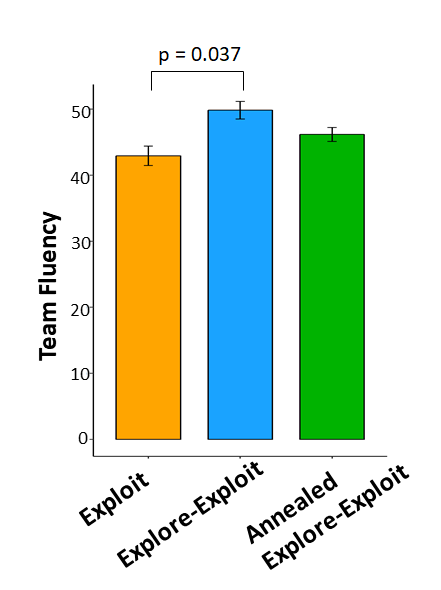}
  \caption{Comparison of Makespan Improvement and Team Fluency across different strategies.}
\label{fig:fluency} 
\end{figure}

\subsubsection{Analysis of H4 - Task Difficulty}
Task difficulty is also assessed using a 7-point Likert scale of the task difficulty questionnaire from \cite{Chao}. We again consider the same linear mixed effects model, with task difficulty as the dependent variable. Upon verifying the assumptions (Shapiro-Wilk's shows $p = 0.255$, and Levene's test shows $p = 0.128$) and performing ANOVA, we did not obtain statistical significance across scheduling strategies ($p=0.2182$). This result complies with our hypothesis that there exists no significant difference in perceived difficulty of the tasks across strategies. 


\section{Discussion}
\label{sec:human_subject_discussion}

Our analysis of H1 indicates there exists significant improvement in schedule makespan for explore-exploit or pure exploit variations of our algorithm. Trends indicate that exploit and explore-exploit have shorter probabilistic distributions for the makespan compared to annealed explore-exploit. Results from H2 show that strategies that include a factor of exploration engenders greater trust, when compared to exploit. Likewise, results from H3 show significantly lower team fluency for pure exploitation. In addition, there is no significant difference in schedule difficulty between the strategies. 

\section{Conclusion}
\label{sec:conclusion}

We conclude that our team coordination algorithm can reason about the future task duration for each individual agent's assignments by improving a model of that individual's learning curve, and use Gaussian upper-bounds to compute a fast evaluation of schedule robustness and makespan upper-bound. Combined, these approaches enable an online genetic algorithm that can evaluate schedules quickly and change assignments in real time to adapt to the needs and improve the efficacy of human-robot teams. Results from human subjects experiment support scheduling strategies that are inclusive of exploration for task allocation in human-robot teams as it engenders greater trust and team fluency, while also improving the makespan.




\bibliographystyle{plainnat}
\bibliography{RSS_20_Submission}

\clearpage

\section{Supplementary}

\subsection{Optimization Workflow}

\begin{algorithm}[!htb]
\SetAlgoLined
 agent\_parameters $\leftarrow $ Compute(population\_priors) \;
 best\_schedules $\leftarrow $ GreedyHeuristic(agent\_parameters) \;
\For{N iterations of scheduling}{
    \While {time $\leq$ Time Limit} { \
    candidate\_schedules $\leftarrow $ SearchMethod(best\_schedules) \;
    best\_schedules $\leftarrow $ UpperBoundEvaluator(candidate\_schedules) \;
    }
    best\_schedule $\leftarrow $ best\_schedules[0] \;
    new\_task\_times $\leftarrow $ Apply(best\_schedule) \;
    agent\_parameters $\leftarrow $ Update(new\_task\_times) \;
}
 \label{workflow_pseudocode}
 \caption{Optimization workflow pseudocode}
 \end{algorithm}

\subsection{Questionnaire} 

\begin{table}[!htb]
\begin{tabular}{|l|}
\hline
\textbf{Post-trial Questionnaire} \\
\hline
\textit{Robot teammate traits} \\
\hline
1. The robot was intelligent. \\
2. The robot was trustworthy. \\
3. The robot was committed to the task. \\
\hline
\textit{Working alliance for human-robot teams} \\
\hline
4. I feel uncomfortable with the robot (reverse scale). \\ 
5. The robot and I understand each other. \\
6. I believe the robot likes me. \\
7. The robot and I respect each other. \\
8. I feel that the robot appreciates me. \\
9. The robot and I trust each other. \\
10. The robot worker perceives accurately what my goals are. \\ 
11. The robot worker does not understand what I am trying to accomplish. \\
12. The robot and I are working towards mutually agreed upon goals. \\
13. I find what I am doing with the robot confusing (reverse scale). \\
\hline
\textit{Task difficulty} \\
\hline
14. The task would be difficult for me to complete alone. \\
15.  The task would be difficult for Sawyer to complete alone. \\
16. The task was difficult for us to complete together. \\
17. My performance was important for completing the task. \\
18. Sawyer’s performance was important for completing the tasks. \\
\hline
\end{tabular}
\label{questionnaire}
\caption{}
\end{table}

\end{document}